\documentclass{ifacconf}

\usepackage{graphicx}      
\usepackage{natbib}       
\usepackage{xcolor}
\usepackage{soul}

\makeatletter

\makeatletter
\let\old@ssect\@ssect 
\makeatother

\usepackage[hidelinks]{hyperref}
\usepackage{xurl}

\makeatletter
\def\@ssect#1#2#3#4#5#6{%
  \NR@gettitle{#6}
  \old@ssect{#1}{#2}{#3}{#4}{#5}{#6}
}
\makeatother

\usepackage{array,ragged2e}
\newcolumntype{P}[1]{>{\RaggedRight\arraybackslash}p{#1}}
\usepackage{booktabs}
\usepackage{tabularx}

\usepackage{amsmath}
\usepackage{amssymb}
\usepackage{amsfonts}

\usepackage{caption}
\usepackage{subcaption}

\newlength\imagewidth
\newlength\imagescale

\usepackage{pgfplots}
\usepgfplotslibrary{external}
\usetikzlibrary{external,calc,patterns,decorations.pathmorphing,decorations.markings,decorations.text,arrows,shapes,positioning, backgrounds}
\usetikzlibrary{arrows.meta,
                backgrounds,
                fit,
                matrix}
\definecolor{ffqqtt}{rgb}{1,0,0.2}
\definecolor{qqqqff}{rgb}{0,0,1}
\pgfplotsset{width=10cm,compat=1.9}

\makeatletter
\renewcommand*\env@matrix[1][\arraystretch]{%
  \edef\arraystretch{#1}%
  \hskip -\arraycolsep
  \let\@ifnextchar\new@ifnextchar
  \array{*\c@MaxMatrixCols c}}
\makeatother

\begin{document}
\begin{frontmatter}

\title{Autonomy for Ferries and Harbour Buses: a Collision Avoidance Perspective} 

\thanks[footnoteinfo]{This research was sponsored by Innovation Fund Denmark, The Danish Maritime Fund, Orients Fund, and the Lauritzen Foundation through the Autonomy part of the ShippingLab project, grant number 8090-00063B. Electronic navigational charts were kindly provided by the Danish Geodata Agency.}

\author[First]{Thomas T. Enevoldsen} 
\author[First]{Mogens Blanke} 
\author[First]{Roberto Galeazzi}

\address[First]{Automation and Control Group, Department of Electrical and Photonics Engineering, Technical University of Denmark, Kgs. Lyngby, DK 2800, Denmark (e-mail: \{tthen,mobl,roga\}@dtu.dk).}

\begin{abstract}                
This paper provides a collision avoidance perspective to maritime autonomy, in the shift towards Maritime Autonomous Surface Ships (MASS). In particular, the paper presents the developments related to the Greenhopper, Denmark's first autonomous harbour bus. The collision and grounding avoidance scheme, called the Short Horizon Planner (SHP), is described and discussed in detail. Furthermore, the required autonomy stack for facilitating safe and rule-compliant collision avoidance is presented. The inherent difficulties related to adhering to the COLREGs are outlined, highlighting some of the operational constraints and challenges within the space of autonomous ferries and harbour buses. Finally, collision and grounding avoidance is demonstrated using a simulation of the whole Greenhopper autonomy stack.
\end{abstract}


\end{frontmatter}

\section{Introduction}
With growing populations and demand of green transitions, compelling needs arise for improved logistic and mobility. Autonomous transportation systems seek to increase efficiency, both in terms of availability, mobility, safety, and emissions. In recent times, rapid development has occurred within maritime autonomy, with the first technological benefits emerging. In the beginning of 2022, the Japanese project, MEGURI2040, demonstrated autonomous capabilities onboard a container vessel. During fall 2021, Sea Machines launched a 1000nm autonomous voyage for their 11m long craft, showcasing the maturity of their technology in inner coastal waters. In fall 2022, the Norwegian autonomous ferry, Milliampere, was demonstrated, highlighting an important case study for inner-city mobility \citep{brekke2022milliampere}. 
\begin{figure}[b]
     \centering
     \begin{subfigure}[b]{1\columnwidth}
            \pgfmathsetlength{\imagewidth}{\linewidth}%
    \pgfmathsetlength{\imagescale}{\imagewidth/524}%
    \begin{tikzpicture}[x=\imagescale,y=-\imagescale]
        \node[anchor=north west] at (0,0) {\includegraphics[trim={20cm 18cm 0cm 4cm},clip,width=0.65\columnwidth]{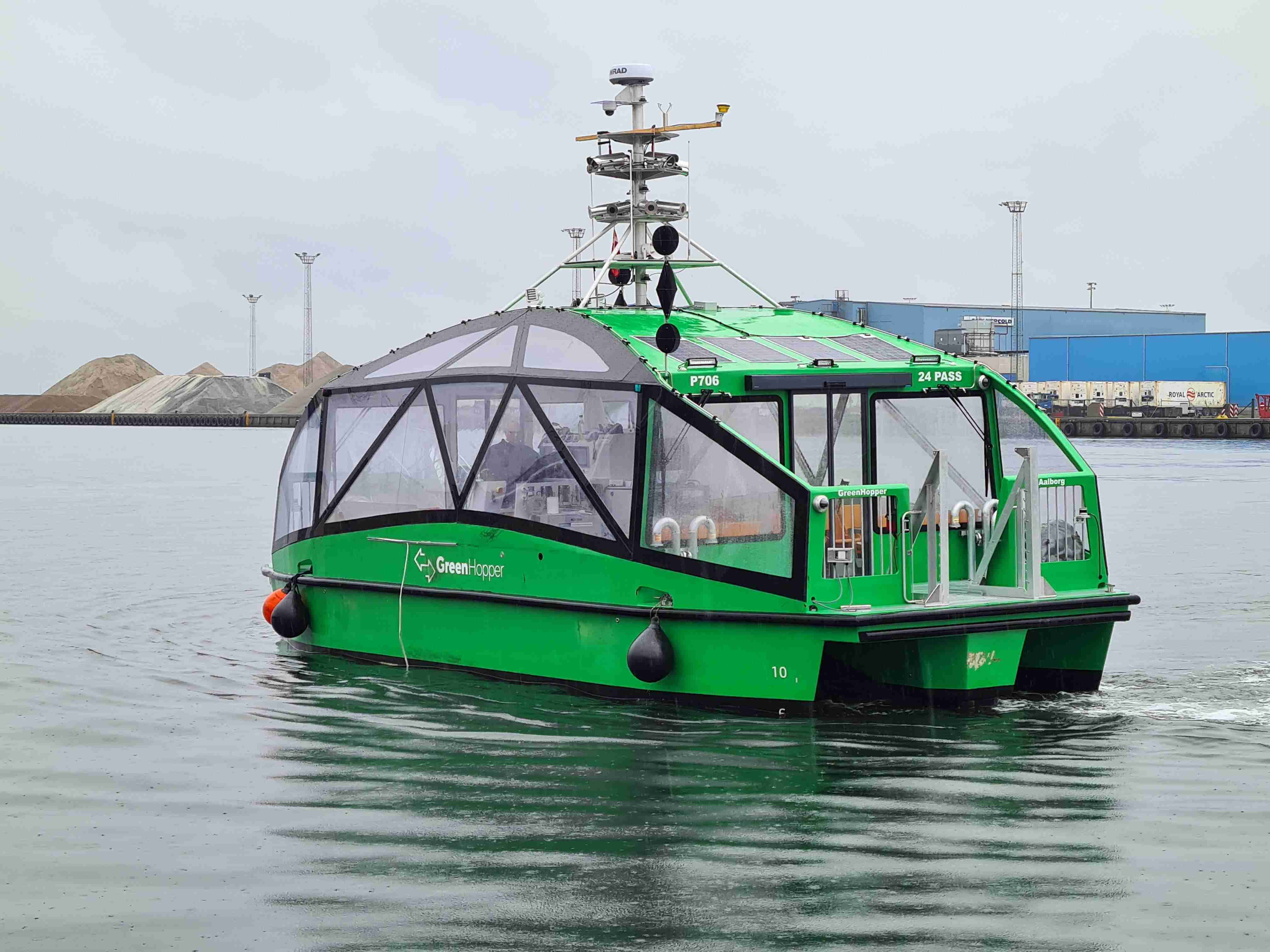}};
        \node[anchor=north west] at (350,0) {\includegraphics[trim={5cm 6cm 6cm 1cm},clip,width=0.3\columnwidth]{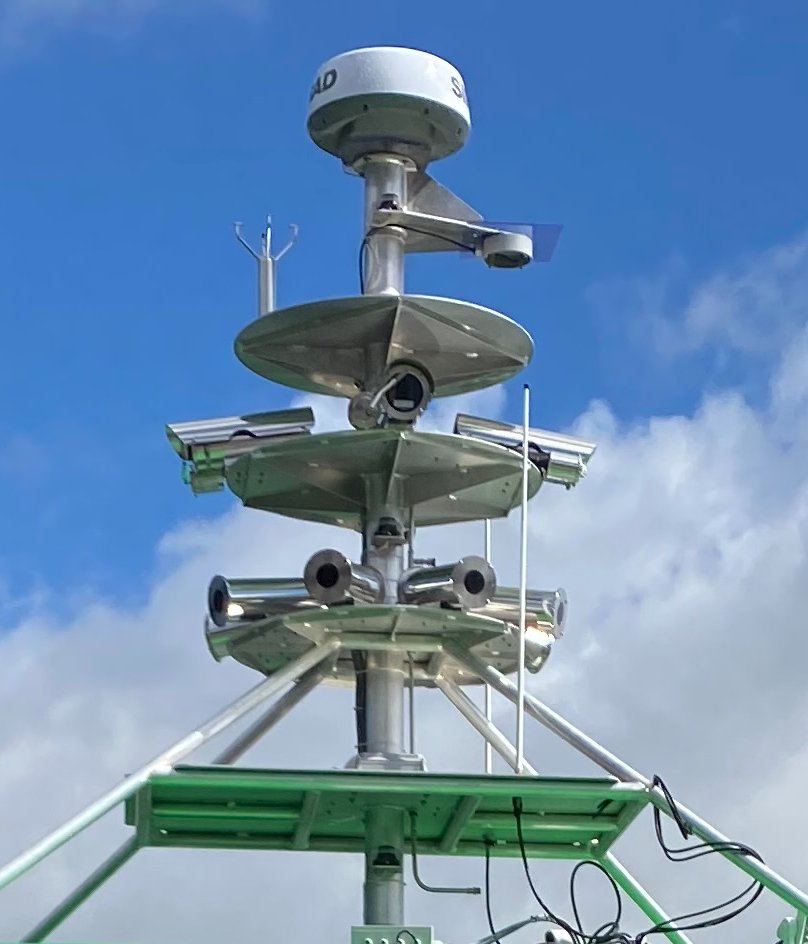}};
    \end{tikzpicture}
	\caption{The Greenhopper: autonomous ferry built by TUCO Yard (DK).}\label{fig:greenhopper}
     \end{subfigure}
      \begin{subfigure}[b]{1\columnwidth}
	\centering
        \includegraphics[width=0.8\columnwidth]{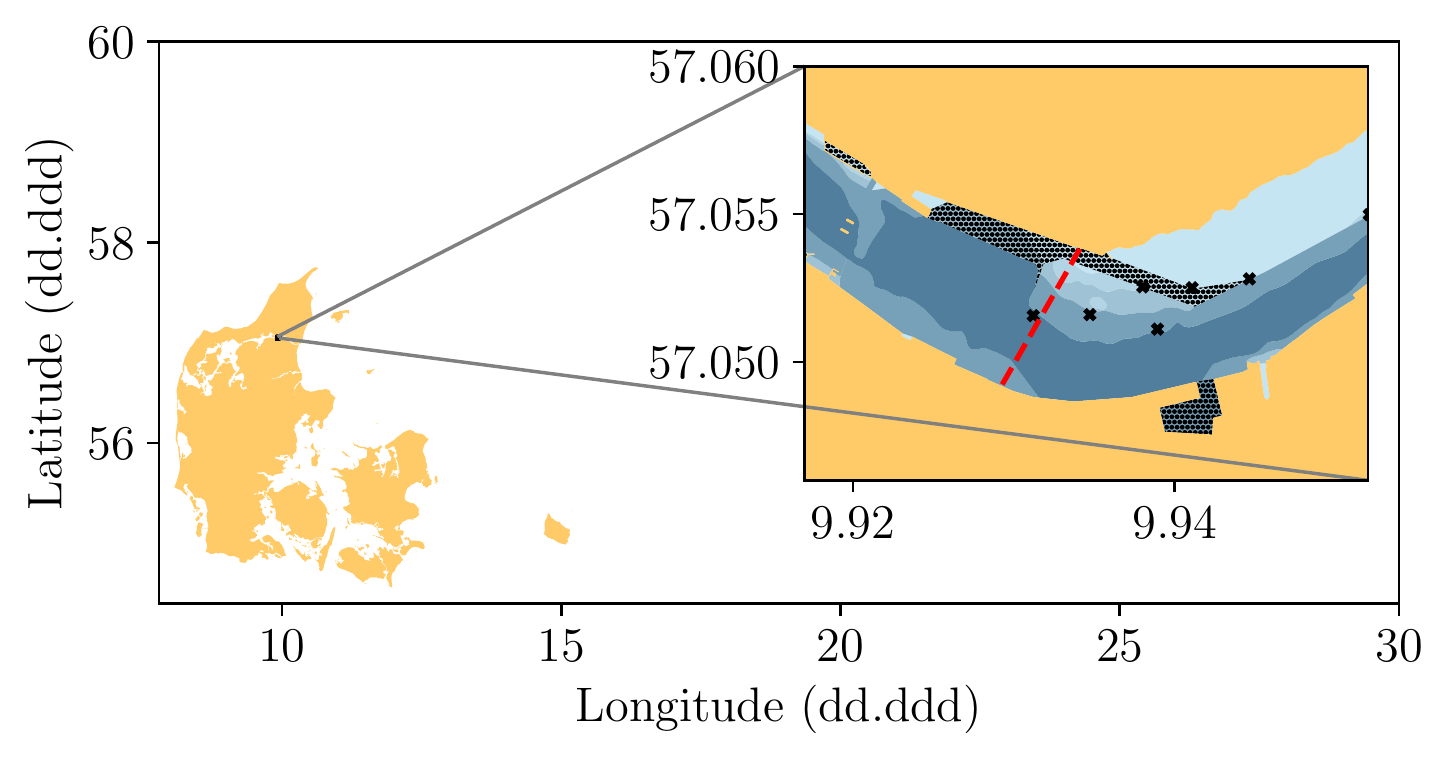}
	\caption{Area of operation at Limfjorden, Aalborg, Denmark. Dashed red line is the nominal route, crosses indicate buoys, hatched areas are dredged and blue shades (light to dark) show increasing water depth.}\label{fig:greenhopper_case_study}
     \end{subfigure}
        \caption{The Greenhopper vessel and its area of operation.}
        \label{fig:case_study}
\end{figure}
By opening waterways for transport of goods and people, Maritime Autonomous Surface Ships (MASS) are prominent candidates to meet increasing mobility demands, strengthen connectivity to island societies, and reduce road congestion in major urban areas. MASS will extend the availability of existing waterborne transportation services, and will open for new mobility on demand services. A major challenge in the shift towards autonomously navigating vessels is the adherence to rules, regulations, and practises laid out by the preceding sailors and navigators.

ShippingLab is a Danish initiative in waterborne transport, where the goal is to create Denmark's first autonomous and environmentally friendly ship. The test and validation case is the Greenhopper, a $12.2$m long double-ended catamaran, driven by electric propulsion. The vessel will facilitate the expansion and growth of the city of Aalborg, located in the northern part of the Danish peninsula, Jutland. It will cross the Limfjorden, a journey of $600$ m.

Recent efforts within collision avoidance for marine autonomy focus on confined and inner coastal waters. In these waters, there are various endeavours that involve computing trajectories in compliance with COLREGs 8 \& 13-17. \cite{bergman2020colregs} demonstrated a two-step optimisation procedure, where a lattice-based planner computes suboptimal trajectories based on motion primitives that are refined by solving optimal control problems (OCP). \cite{enevoldsen2022sampling} presented a sampling-based method to calculate minimal route deviations, minimising cross-track error and speed loss. \cite{thyri2022field} detailed a collision avoidance scheme that assigns and uses control barrier functions for preventing ship domain violation, and thereby enforcing the COLREGs. 

For the MilliAmpere \citep{brekke2022milliampere}, \cite{bitar2021three} detailed a method consisting of the three aspects of an autonomous voyage: undocking, transit, and docking. Docking was dealt with using model predictive control, whereas the transit phase combined a hybrid A* with an OCP solver. Leveraging a set of pre-defined feasible paths, \cite{thyri2020path} instead cast the problem as velocity planning. The planning phase recomputed with respect to dynamic obstacles.
The Dutch project Roboat sought to implement an autonomous platform for urban mobility \citep{wang2020roboat}, where in \citep{de2022regulations} the system demonstrated its capabilities and basic adherence to COLREGs rules 13-15. 
For the Rhine river, \cite{Koschorrek2022} presented a system that used a hybrid A* to find feasible trajectories. COLREGs were not directly considered because local law dictates that a ferry must yield for any other traffic.

This paper presents a collision avoidance scheme, the \textit{Short Horizon Planner} (SHP), designed for a fjord crossing ferry, the Greenhopper. The SHP considers the available manoeuvrability for precise obstacle avoidance, while partially adhering to the IMO COLREGs (rules 8 \& 13-17). The role, purpose, and responsibility of the collision avoidance system within the autonomy stack is detailed and discussed, outlining apparent operational constraints in both the collision avoidance system and the remaining stack. The particular operation of the Greenhopper is detailed, highlighting the interplay between the SHP and the remaining autonomy components.

\begin{figure}[t]
\centering
\begin{subfigure}[b]{1\columnwidth}
	\includegraphics[width=1\columnwidth]{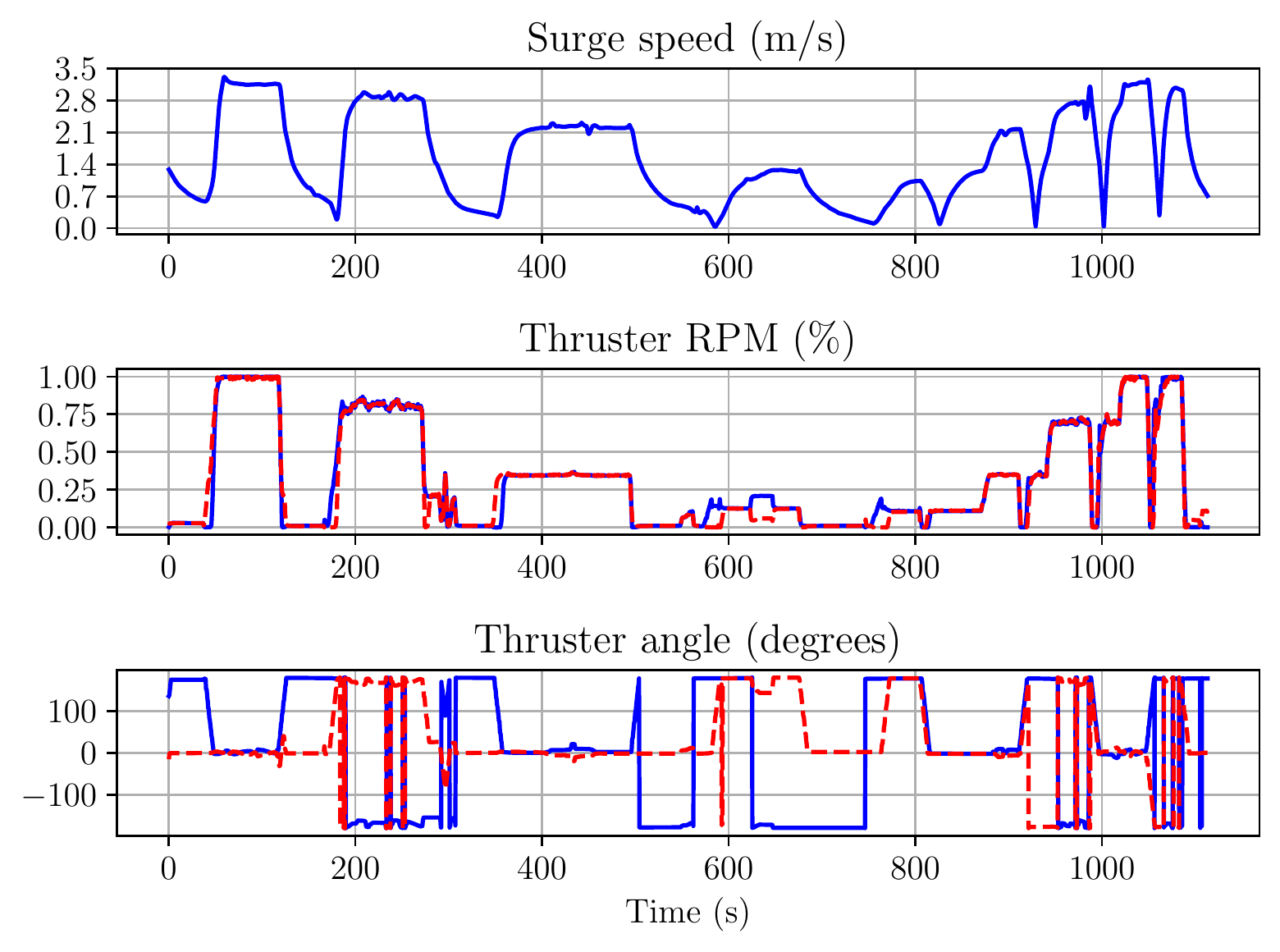}
	\caption{Experimental data from the Greenhopper. $t=0s$ to $t=~800s$ contains acceleration and deceleration experiments. From $t=800s$ and onwards are recorded emergency stops at various speeds.}\label{fig:greenhopper_exp}
\end{subfigure}
\hfill
\begin{subfigure}[b]{0.49\columnwidth}
	\centering
\includegraphics[width=1\columnwidth]{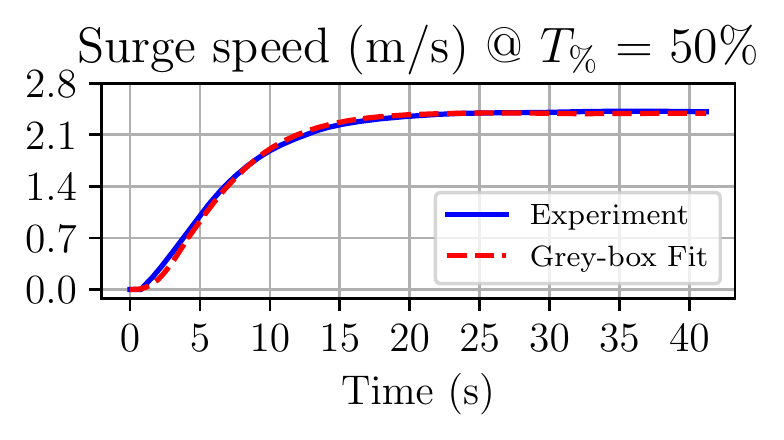}
\caption{Grey-box estimate of \eqref{eq:greyboxmodel} at 50\% thrust.}\label{fig:greenhopper_SYSID}
\end{subfigure}
\hfill
\begin{subfigure}[b]{0.49\columnwidth}
	\centering
\includegraphics[width=1\columnwidth]{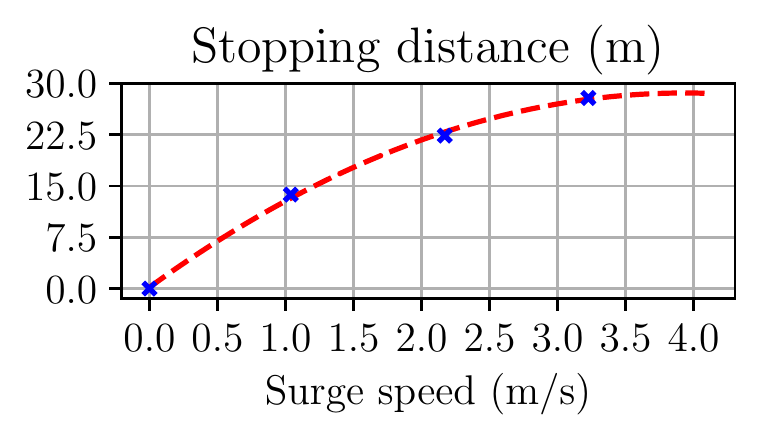}
\caption{Second order fit for the emergency stopping distances.}\label{fig:greenhopper_curvefit}
\end{subfigure}
\caption{Experiments and estimates from the Greenhopper.}\label{fig:greenhopper_data}
\end{figure}

\section{System modelling and identification}
The Greenhopper is propelled and manoeuvred by two azimuth thrusters, located fore and aft, at the centre line. It is equipped with four RGB cameras, eight Long Wavelength Infrared (LWIR) cameras, four W band and one X band radar, two 3D lidars, a GNSS, a gyro compass, an AIS transponder and an IMU. Sensors mounted on the mast can be seen in Fig.~\ref{fig:greenhopper}. A Voyage Control System (VCS) is responsible for executing steering and track control along a nominal route. The VCS will safely dock, undock, and carry out the voyage in nominal conditions. The nominal route can be modified by adding supplemental waypoints to the VCS, as safe navigation requires.

\subsection{Surge velocity dynamics}
Surge acceleration is the result of the balance between propeller forces and hull resistance.
With azimuth thrusters fore and aft, along ship thrust is, 
\begin{equation}
    T_x = T_{P,1}\cos(\phi_1) + T_{P,2}\cos(\phi_2)
\end{equation}
where $\phi_i$ is the azimuth angle of thruster $i$ and $T_{P,i}$ is the propeller thrust.

Hull resistance at low Froude number consists of Stokes friction, linear in $u$, and pressure drag, proportional to $u|u|$. With mass and added mass in the left hand side factor, and thrust deduction $t$, the main components in the surge dynamics are,
\begin{equation}\label{eqn:surgeacc}
    (m - X_{\dot{u}})\dot{u} = (1-t)T_x + X_{u}u + X_{u|u|} u|u| + (m+X_{vr})vr
\end{equation}
where $v$ is sway velocity and $r$ is turning rate in body coordinates. The term $vr$ represents fictive acceleration in the rotating body. In steady state, the three resistance terms on the right hand side balance the effective along-ship's thrust $(1-t)T_x$. 
Introducing $T_x = \beta_s T_\%$ with $T_\%$ being commanded thrust percentage, \eqref{eqn:surgeacc} has the form, 
\begin{equation}\label{eq:greyboxmodel}
    \dot{u} = \beta T_\% + \alpha u + \gamma u|u| + \delta vr
\end{equation}
with scaled parameters for propeller thrust $\beta>0$, linear damping $\alpha<0$, quadratic damping $\gamma<0$ and fictive acceleration $\delta<0$, respectively. The scaled parameters are identified from full scale testing in the following section.

\subsection{Grey-box identification and analytical solution}
The identification of \eqref{eq:greyboxmodel} based on full-scale straight line acceleration data revealed that, due to the low speed regime, the particular hull form, and the propeller slip stream interaction with the pontoons of the catamaran hull, the nonlinear damping coefficients $\gamma$ and $\delta$ do not contribute at low speeds. Therefore, a linear dynamical model is identified based on available experimental data using the prediction error method, 
\begin{equation}\label{eq:linear_motion_model}
    \dot{u} = \beta T_\% + \alpha u, \quad \dot{N} = \cos\left(\psi\right) u, \quad \dot{E} = \sin\left(\psi\right) u
\end{equation}
where $T_{\%}$ and heading angle $\psi$ are specified for each segment of the path. The solution of \eqref{eq:linear_motion_model}
\begin{equation}\label{eq:linear_solution}
    \begin{bmatrix}[1.5]
        u(t)\\ N(t) \\ E(t)
    \end{bmatrix} = 
    \begin{bmatrix}[1.5]
        \left(1 - e^{\alpha t}\right)\frac{\beta}{-\alpha} T_{\%} \\
        \left(\frac{1}{\alpha^2}e^{\alpha t} - \frac{1}{\alpha}t\right)T_{\%}\beta\cos(\psi)\\
        \left(\frac{1}{\alpha^2}e^{\alpha t} - \frac{1}{\alpha}t\right)T_{\%}\beta\sin(\psi)
    \end{bmatrix} . 
\end{equation}
The analytical solution is used to calculate a trajectory between two points in the north-east plane, by computing arrival time $t_f$ at the final point. Arrival time is obtained by setting the left-hand side of \eqref{eq:linear_solution} to the desired point and solving for $t$. Obtaining $t_f$ for all segments yields a combined trajectory, apart from short transient phases when passing way-points.

\section{Path Planning for collision avoidance}\label{sec:planner}
The nominal path of the Greenhopper is described by fixed waypoints in the VCS. In conditions with traffic, the objective of the SHP is to find a sequence of states that connects either the nominal waypoints, or the current position of ownship and the goal in a collision-free and safe manner.

\subsection{Spatio-temporal lattice planner}\label{sec:lattice_planner}
Let $\mathcal{X} \subseteq \mathbb{R}^{2}\times\mathbb{R}_{\geq0}$ be the state space, with $\mathrm{x} \in \mathcal{X}$ and $\mathrm{x} = [E, N, t]^T$. $\mathcal{X}$ is divided into two subsets, the free space $\mathcal{X}_{\text{free}}$ and the obstacle space $\mathcal{X}_{\text {obs}}$, with $\mathcal{X}_{\text{free}}=\mathcal{X} \backslash \mathcal{X}_{\text{obs}}$. The objective is to find a time-monotonic sequence $\sigma$ of states that minimises the cost function $c(\sigma)$, while connecting the start $\mathrm{x}_{\text{s}}$ and end $\mathrm{x}_{\text{e}}$ states
\begin{equation}
\begin{split}
\sigma^{*}=\underset{\sigma \in \Sigma}{\arg \min }\left\{c(\sigma) \mid \right. \sigma(0)=\mathrm{x}_{\text {s }},\, &\sigma(1)=\mathrm{x}_{\text {e }}, \\
\forall s \in[0,1],\, &\sigma(s) \in \left. \mathcal{X}_{\text {free }}\right\}.
\end{split}
\end{equation}
The obstacle subset $\mathcal{X}_{\text{obs}}$ is formed by the union over all constraints \citep{enevoldsen2022sampling}, namely
\begin{equation}
    \mathcal{X}_{\text{obs}} = \mathcal{X}_{\text{obs}}^{\text{OS}} \cup \mathcal{X}_{\text{obs}}^{\text{ENC}} \cup \mathcal{X}_{\text{obs}}^{\text{TV}}, \quad\quad\, \mathcal{X}_{\text{obs}}^{\text{TV}} = \bigcup^n_{i=1}  \mathcal{X}_{\text{TV},i}
\end{equation}
with $\mathcal{X}_{\text{obs}}^{\text{OS}}$ containing states that violate manoeuvring constraints, $\mathcal{X}_{\text{obs}}^{\text{ENC}}$ the grounding and buoy collision states, and finally $\mathcal{X}_{\text{obs}}^{\text{TV}}$ the target vessel constraints, which is the union of $n$ vessels, such that all $n$ are considered simultaneously. The spatial constraints are encoded by the predicted trajectories of each target vessel ($\mathcal{X}_{\text{TV},i}$). 

\begin{figure}[t]
\centering
\begin{subfigure}[b]{1\columnwidth}
	\centering
        \includegraphics[width=0.82\columnwidth]{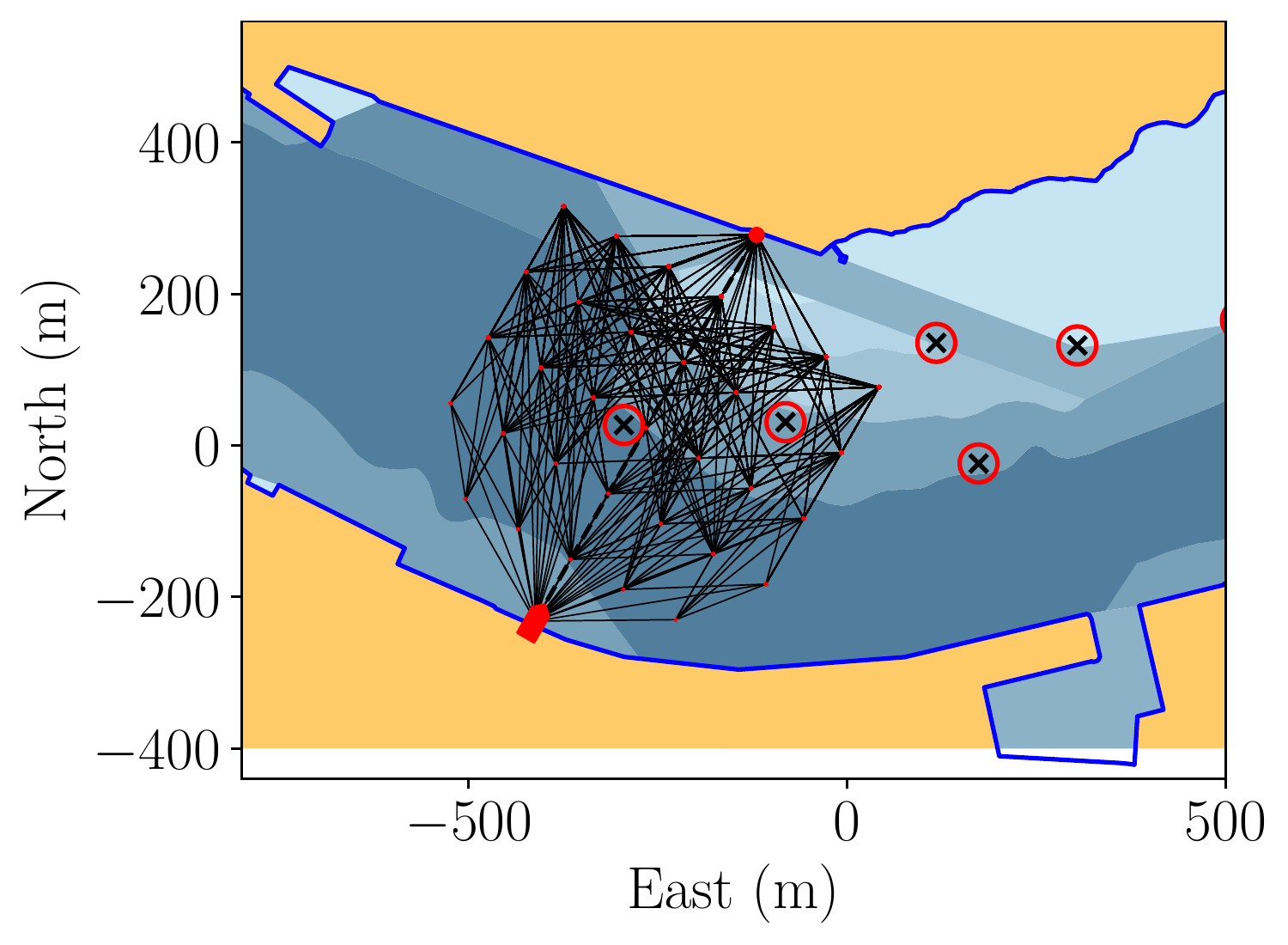}
	\caption{Full lattice using a 7x5 grid.}\label{scenario_limfjorden}
\end{subfigure}
\hfill
\begin{subfigure}[b]{1\columnwidth}
	\centering
        \includegraphics[width=0.82\columnwidth]{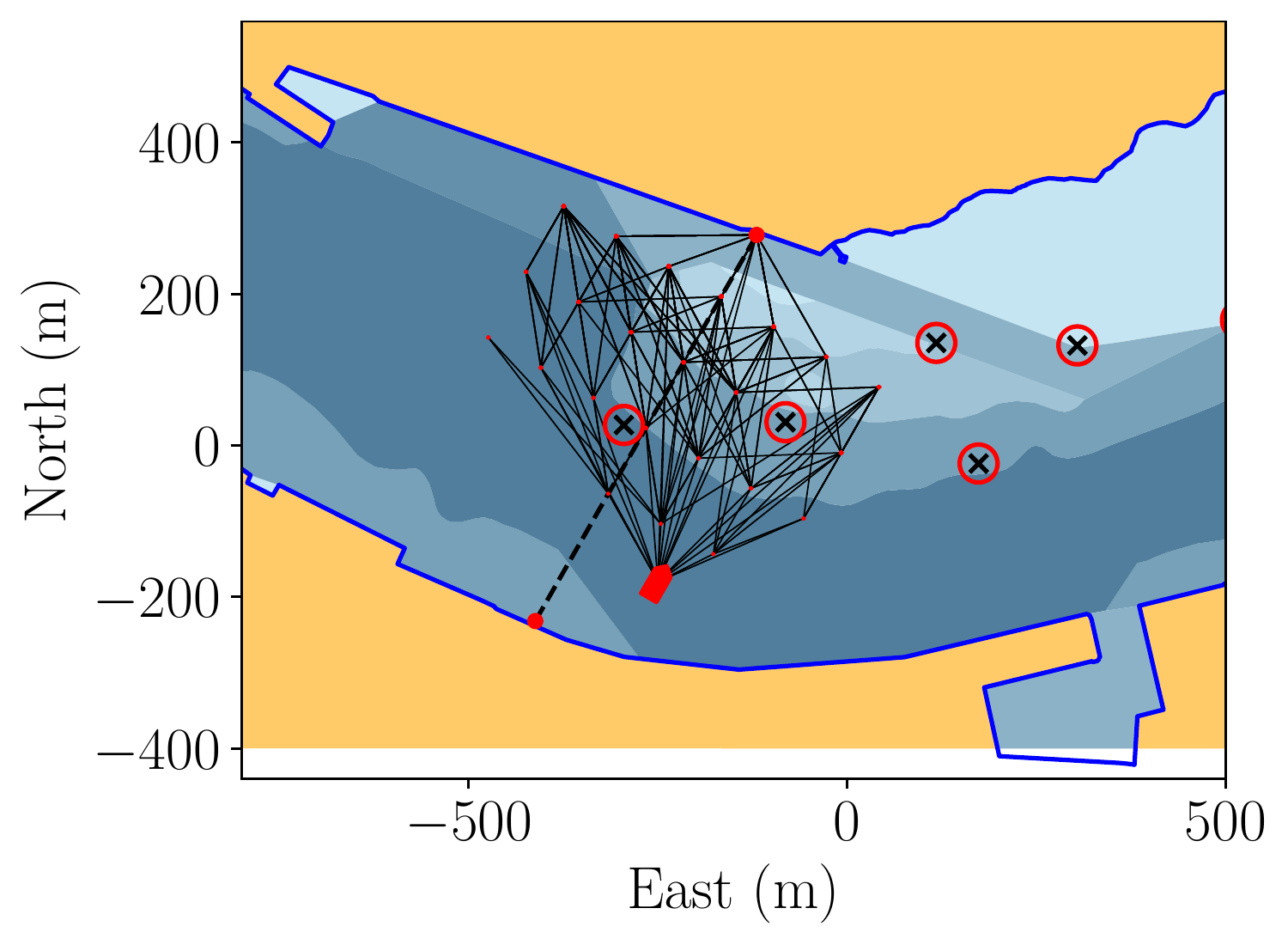}
	\caption{Recomputing the lattice from an arbitrary location.}\label{scenario_limfjorden_voyage}
\end{subfigure}
\caption{The obstacle subset $\mathcal{X}_{\text{obs}}^{\text{ENC}}$ consists of the area surrounding the blue polygon (land and shallow waters) and the interior area of the red circles (buoys).}\label{fig:lattice_planner_example}
\end{figure}
A deterministic planning algorithm is proposed for building a directed graph. The starting state $\mathrm{x}_{\text{s}}$ is the position of ownship in the north-east plane at $t=t_0$, and $\mathrm{x}_{\text{e}}$ is the desired destination, at a unknown final time $t = t_f$. Consider a grid $\mathcal{G} = [\mathrm{G}_1, \ldots, \mathrm{G}_m]^T \in \mathbb{R}^{m \times n}$, with rows $\mathrm{G}_i = [g_{i,1} \ldots, g_{i,n}]$ , where each row represents the depth towards $\mathrm{x}_e$ and the width of potential deviations, and each element $g_{i,j} \in \mathcal{G}$ represents a point in the north-east plane. 
A refined grid, $\bar{\mathcal{G}} = \left[\bar{\mathrm{G}}_1, \ldots, \bar{\mathrm{G}}_m\right]^T$, is obtained by removing from $\mathcal{G}$ elements that violate $\mathcal{X}_{\text{obs}}^{\text{ENC}}$, i.e. $\forall\, g_{i,j} \notin \mathcal{X}_{\text{obs}}^{\text{ENC}}, \, \bar{g}_{i,j} = g_{i,j}$.

A directed graph $\mathcal{T}$ with root $\mathrm{x}_{\text{s}}$ is built over $M = m+1$ iterations. Two sets of nodes are formed, one with all the currently considered parent nodes $\mathcal{C}_p = \{\mathrm{x}_{\text{s}}\}$, which always contains $\mathrm{x}_{\text{s}}$, and a set for all the currently considered child nodes $\mathcal{C}_c = \{\mathrm{x}_{\text{e}}\}$ that always contains $\mathrm{x}_{\text{e}}$. Each iteration the sets are modified by the given grid rows, which dictates edges that are to be formed between the nodes in each set
\begin{align}
	&\mathcal{C}_p = \{\mathrm{x}_{\text{s}}\},\, \mathcal{C}_c = \{\mathrm{x}_{\text{e}}, \bar{\mathrm{G}}_k, \bar{\mathrm{G}}_{k+1},\dots, \bar{\mathrm{G}}_m\} &&\text{if } k = 1, \nonumber \\
	&\mathcal{C}_p = \{\mathrm{x}_{\text{s}}, \bar{\mathrm{G}}_{k-1}\},\, && \nonumber\\
        &\quad\quad\quad\quad\mathcal{C}_c = \{\mathrm{x}_{\text{e}}, \bar{\mathrm{G}}_k, \bar{\mathrm{G}}_{k+1},\dots, \bar{\mathrm{G}}_m\} &&\text{if } 1 < k <M, \nonumber\\
	&\mathcal{C}_p = \{\mathrm{x}_{\text{s}}, \bar{\mathrm{G}}_{k-1}\},\, \mathcal{C}_c = \{\mathrm{x}_{\text{e}}\} &&\text{if } k = M . \nonumber
\end{align}
Before adding a given edge to $\mathcal{T}$, the resulting trajectory between the two nodes is checked to see if it violates any constraints ($\mathcal{X}_{\text{obs}}$). The trajectory between two nodes is computed by forward simulating \eqref{eq:linear_solution}. Nodes in collision from $\mathcal{C}_c$ are discarded and omitted from $\mathcal{C}_p$.

\subsection{Rules and regulations}
Adherence to the rules and practises of safe navigation is fundamental for MASS. There is a general consensus that the most essential IMO COLREGs are rules 8 \& 13-17. These apply to common vessel encounters, namely: overtaking (13), head-on (14) and crossing (15). Rule 14 \& 15 specify that the give-way vessel must perform a manoeuvre toward starboard, rule 13 allows passing on either side in a safe manner. Rule 16 \& 17 dictate the behaviour of the give-way and stand-on vessel, see \citep{cockcroft2003guide}. In the literature, there is a consensus that partial adherence to the COLREGs is sufficient to demonstrate capable and safe navigation. Within confined waters, however, such as rivers and urban environments, additional complexities may arise. Rule 14 and 15 specifically apply between two power-driven vessels; therefore, if the system encounters a sailboat, different rules and obligations apply. Furthermore, certain vessels may be restricted in their manoeuvrability. If a vessel is restricted, rule 9 applies, which states that a vessel less than 20m should give-way for a restricted vessel, even if according to rule 15 the target vessel is the give-way vessel. \cite{hansen2022autonomous} used rule 9 within the decision making loop for a river crossing ferry, emphasising that following rule 15 without considering rule 9 may cause problems. 

The complexity of the rule framework strongly depends on local regulations (e.g. the Rhine River \citep{Koschorrek2022}) for the particular waters. In Canadian waters, rule 15 is modified so that any vessel, with minor exceptions, crossing a river must yield to power-driven vessels travelling along it \citep{CanadianRule15}. According to \S 19 of Danish law on seafaring \citep{SofartsstyrelsenGeneral,SofartsstyrelsenLimf}, only three specific ferry routes must disregard the usual obligation in rule 15, and instead yield for any traffic the ferry may impede. 

\subsection{Ship domains for COLREGs-compliance}\label{sec:ship_domains}
\begin{figure}[t]
\centering
\begin{subfigure}[b]{0.325\columnwidth}
	\centering
        \includegraphics[trim={0cm 0cm 0cm 7cm},clip,width=1\columnwidth]{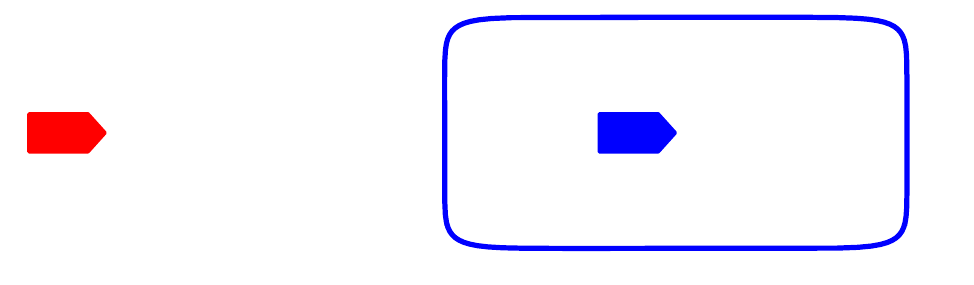}
	\caption{Overtaking (13)}\label{limf_bumper_two_vessels_overtaking}
\end{subfigure}
\hfill
\begin{subfigure}[b]{0.325\columnwidth}
	\centering
        \includegraphics[trim={0cm 0cm 0cm 4cm},clip,width=0.85\columnwidth]{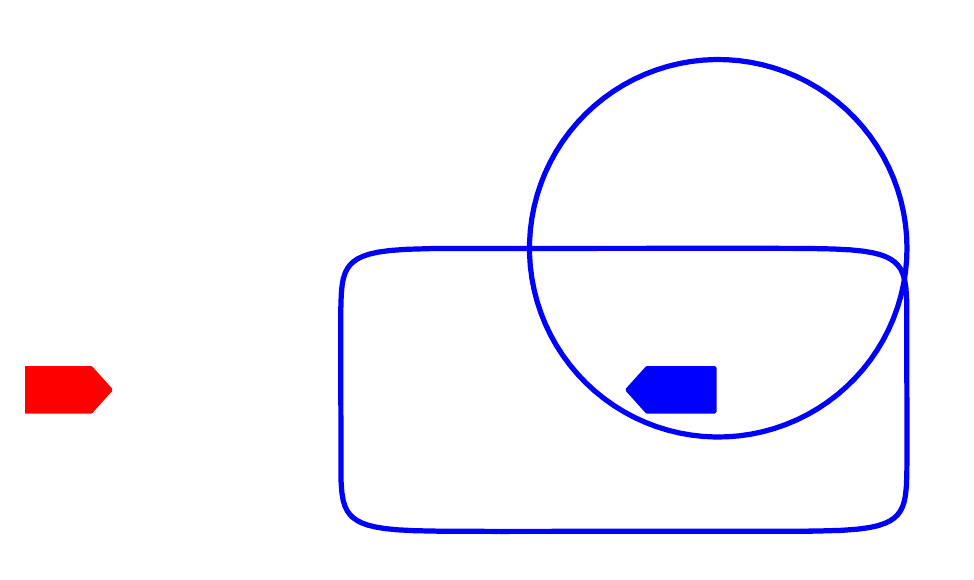}
	\caption{Head-on (14)}\label{limf_bumper_two_vessels_headon}
\end{subfigure}
\hfill
\begin{subfigure}[b]{0.325\columnwidth}
	\centering
        \includegraphics[trim={1cm 2.2cm 0cm 3cm},clip,width=0.9\columnwidth]{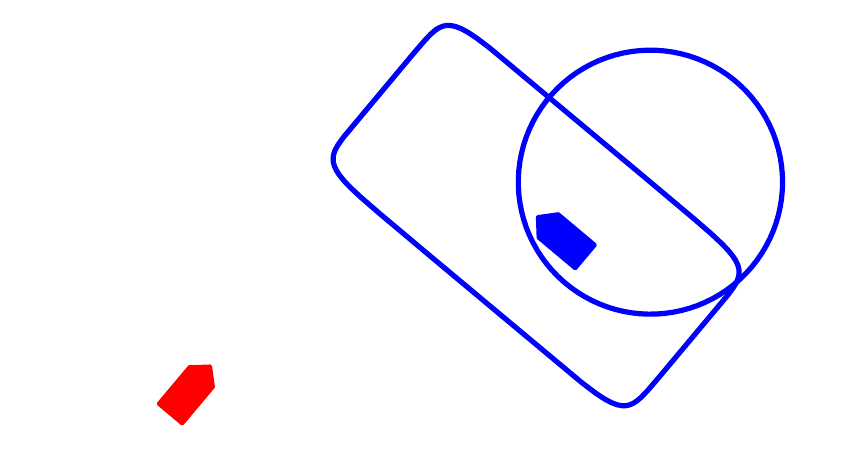}
	\caption{Crossing (15)}\label{limf_bumper_two_vessels_crossing}
\end{subfigure}
\caption{Ship domains for enforcing COLREGs-compliance, dimensions are dependant on target vessel ship length.}\label{fig:limf_bumper}
\end{figure}
If the given MASS has adequate situation awareness, give-way and stand-on obligations can be enforced using specialised ship domains, see Fig.~\ref{fig:limf_bumper}, which discards states in violation with the given COLREGs scenario. Domains for complying with crossing and overtaking scenarios are described by a Lamé curve \citep{enevoldsen2022sampling}
\begin{equation}\label{eq:COLREGs_lame}
    \begin{split} 
    &\left| \frac{\cos(\psi(t))\Delta E_l(t) - \sin(\psi(t))\Delta N_l(t)}  {a_L}\right|^p            +\\
    &\left| \frac{\sin(\psi(t))\Delta E_l(t) + \cos(\psi(t))\Delta N_l(t)}{b_L}\right|^p            \leq 1
    \end{split}
\end{equation}
combined with a circular constraint,
\begin{equation}\label{eq:COLREGs_circle}
   \Delta E_c(t)^2 + \Delta N_c(t)^2 \leq r_L^2
\end{equation}
where $a_L$, $b_L$ and $r_L$ are scalar values based on the length of the target vessel and additional safety margins. The difference in coordinates at time $t$ between the own ship and a given target vessel is used to evaluate the domains $\Delta E(t) = E(t) - E_{\text{TV}}(t) + E_{\text{o}}$, $\Delta N(t) = N(t) - N_{\text{TV}}(t) + N_{\text{o}}$ and $\bar{\psi}(t) = \psi_{\text{TV}}(t) + \psi_{\text{o}}$, where the offset is used to shift the elliptical and circular components of the domain,
\begin{equation}
    \begin{bmatrix}[1.5]
    E_{\text{o}}\\ N_{\text{o}}
    \end{bmatrix}
    = 
    \begin{bmatrix}[1.5]
        \cos\left(-\bar{\psi}\right) & -\sin\left(-\bar{\psi}\right)\\
        \sin\left(-\bar{\psi}\right) & \cos\left(-\bar{\psi}\right)\\
    \end{bmatrix}
    \begin{bmatrix}[1.5]
        p_E\\p_N
    \end{bmatrix}
\end{equation}
with $p_E = 0$, and $p_N$ equal to $b_L$ in \eqref{eq:COLREGs_lame} and $a_L$ in \eqref{eq:COLREGs_circle}. For \eqref{eq:COLREGs_lame} and \eqref{eq:COLREGs_circle}, the $\psi_{\text{o}}$ is equal to $0$ and $\frac{\pi}{2}$ respectively.

\subsection{Safety margins}
Ship length is commonly used to compute a safety margin with respect to other vessels. However, vessels navigating within inner coastal or confined waters are typically either sailboats or pleasure crafts, which are not obligated to carry an AIS transponder. Vessels of length greater than $20$m are often required to have AIS.
Therefore, it is necessary to select an adequate safety margin in the absence of an accurate ship length estimate. The safety margin is selected according to emergency stop manoeuvres (Fig.~\ref{fig:greenhopper_exp}), such that a suitable distance is maintained to the target vessel, in instances with an erroneously perceived scenario. For the current transit speed of 3 knots, the required stopping distance is approximately $17.5$m, therefore selecting $25$m as the minimum ship length is ample distance.

\begin{figure}[tb]
     \centering
        \centering
	\includegraphics[trim={0cm 0.6cm 0cm 1.5cm},clip,width=1\columnwidth]{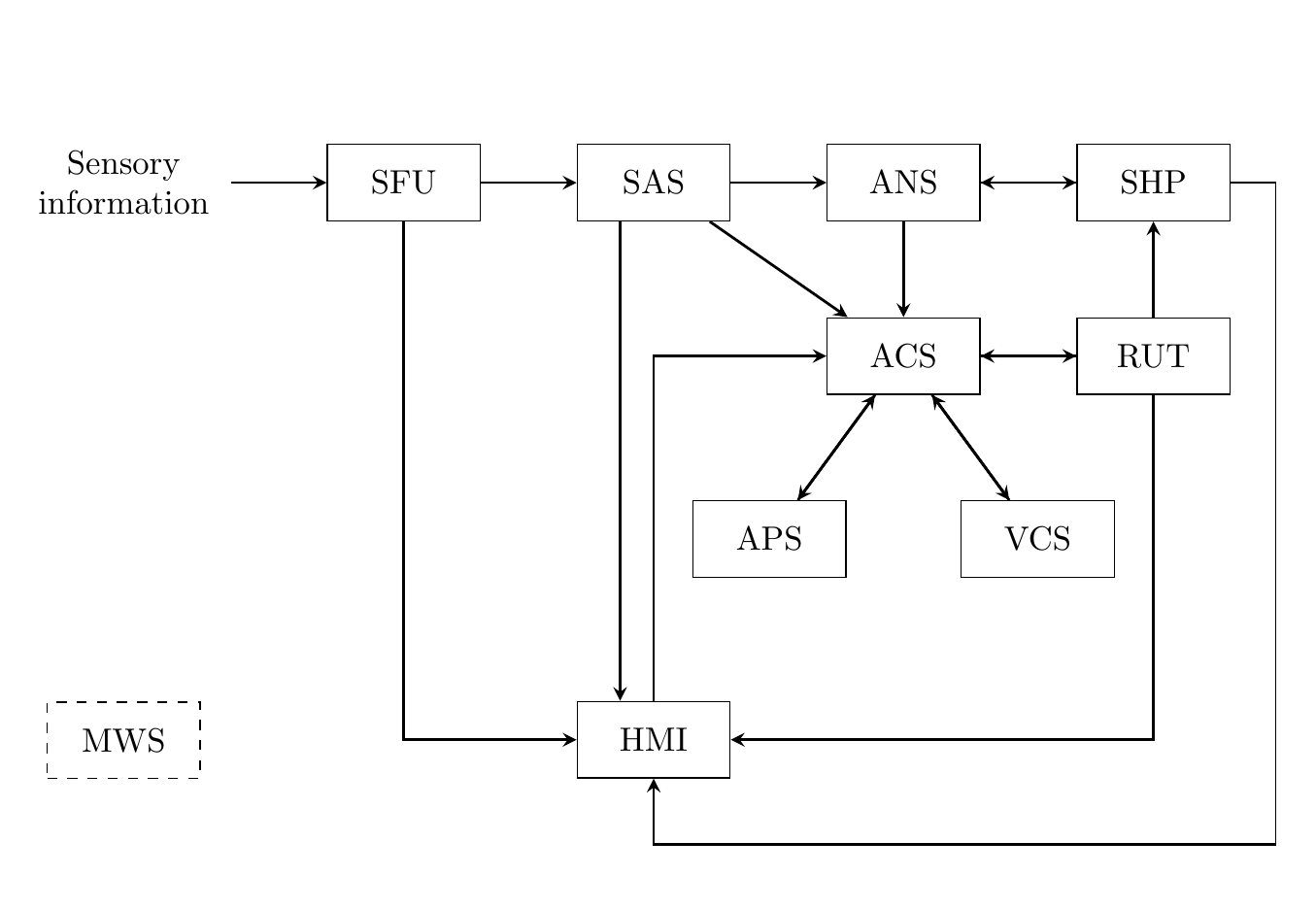}
	\caption{Module interconnection within the autonomy stack.}\label{middleware_modules}
\end{figure}
\section{The autonomy stack}
The following section details the composition of the autonomy stack. The middleware is introduced and the purpose and responsibility of each module is outlined.

\subsection{Middleware and autonomy stack features}
\cite{Dittmann_2022} investigated the regulatory framework and system requirements for the development and commissioning of MASS, highlighting some important considerations regarding reliability and redundancy. In addition, design choices and developments related to the custom middleware solution were detailed.

The autonomous system is composed of various modules, such that each block is compartmentalised and its interface clearly defined. Using a modular approach allows each part of the stack to be developed and tested individually, and also undergo strict stress and acceptance testing, before being rolled out and combined with the remaining system. The middleware facilitates publish–subscribe communication between modules, such that multiple modules can subscribe to the same module. Testing and simulation of various modules within the stack is achieved using a dedicated middleware simulator (MWS).

\subsection{Module functionality and interconnection}
The core of the stack consists of the Autonomous Coordination Supervisor (ACS), Autonomous Navigation Supervisor (ANS) and Autonomous Platform Supervisor (APS), which embodies the traditional roles employed by the captain, navigator, and chief engineer \citep{dittmann2021autonomy}. The ACS coordinates departures and exchanges routes with the route server (RUT), which stores the destination.

Effective and precise fused perception and sensory information is crucial for the remaining autonomous system. Human lookouts and navigators are replaced by an electronic outlook \citep{blanke2018outlook} that uses cameras to detect and classify objects \citep{scholler2020vision}. The vision system is fused with the remaining sensors, producing a robust and resilient estimate of static and dynamic obstacles \citep{Dagdilelis2022}, all of which is encapsulated by the Sensor Fusion (SFU) module. The estimated and fused states of the surrounding vessels can be augmented by a trajectory prediction scheme that uses information from the local area \citep{scholler2021trajectory}. However, the current stack only implements straight-line predictions.
The Situation Awareness Service (SAS) is driven by information from the SFU, in order to maintain an overview of the unfolding scenario. Once a vessel violates chosen Closest
Point of Approach (CPA) and Time to CPA (TCPA), limits, the scenario is passed from SAS to the ANS, i.e. the 'navigator' is informed about the situation, and triggers the SHP for a route deviation. This ensures that the Greenhopper deals with the scenario in a timely manner and with a reasonable safety margin. Details on the interaction between the SAS and ANS modules can be found in \cite{hansen2020DES} and \cite{Papageorgiou_2022}. A Human-machine Interface (HMI) visualises the SFU, RUT, SAS and SHP on an electronic navigational chart, with correct symbolism from IMO. 

The SHP, originally introduced in \cite{enevoldsen2022sampling} as a generalised collision avoidance scheme for vessels in confined and inner coastal waters, is in this paper specialised for crossings, such as those encountered by the Greenhopper. The underlying planning scheme is implemented as described in Section \ref{sec:planner}. As part of the autonomy stack, the SHP is tasked with computing rule-compliant, collision-, and grounding free passages for a given scenario at hand. Given an input consisting of own ship navigational data, predicted target vessel information, perceived COLREGs scenario (from the SFU-SAS-ANS) and the current destination (from RUT). The SHP reports within finite time whether a valid crossing exists and, if so, which sequence of waypoints must be followed to achieve it.  

\begin{figure}[tb]
     \centering
     \begin{subfigure}[b]{0.494\columnwidth}
        	\centering
        \includegraphics[width=0.95\textwidth]{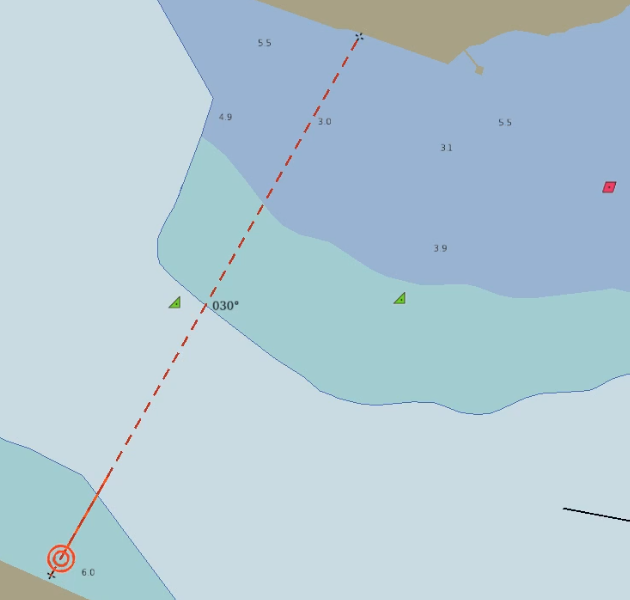}
	\caption{t=10s}\label{fig:t10s}
     \end{subfigure}
     \hfill
       \begin{subfigure}[b]{0.494\columnwidth}
        	\centering
        \includegraphics[width=0.95\textwidth]{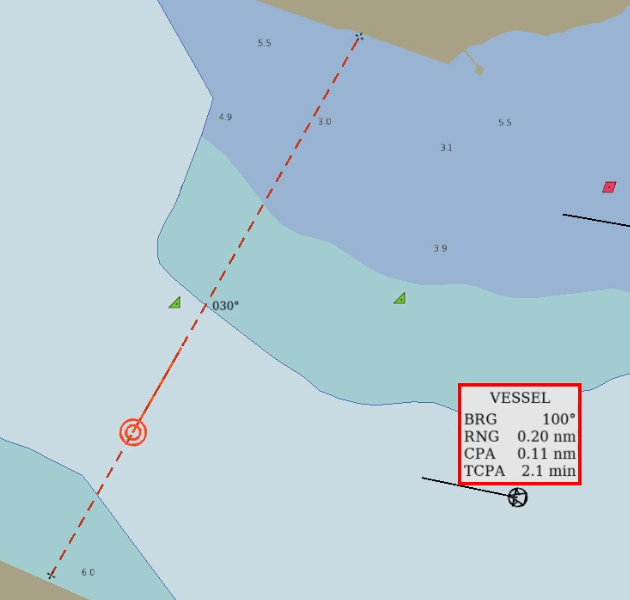}
	\caption{t=1m40s}\label{fig:t1m40s}
     \end{subfigure}
     \hfill
       \begin{subfigure}[b]{0.494\columnwidth}
        	\centering
        \includegraphics[width=0.95\textwidth]{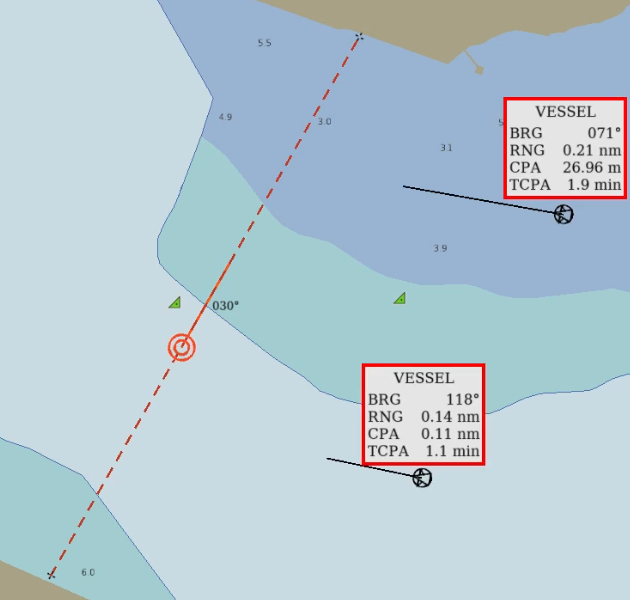}
	\caption{t=2m40s}\label{fig:t2m40s}
     \end{subfigure}
     \hfill
       \begin{subfigure}[b]{0.494\columnwidth}
        	\centering
        \includegraphics[width=0.95\textwidth]{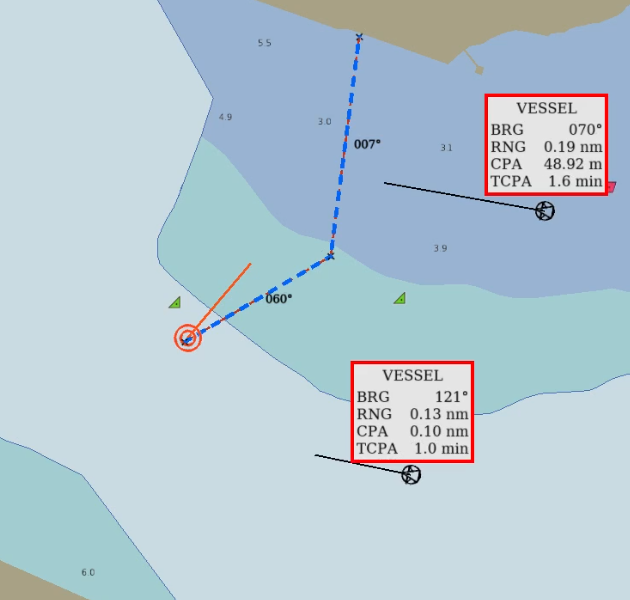}
	\caption{t=2m47s}\label{fig:t2m47s}
     \end{subfigure}
     \hfill
       \begin{subfigure}[b]{0.494\columnwidth}
        	\centering
        \includegraphics[width=0.95\textwidth]{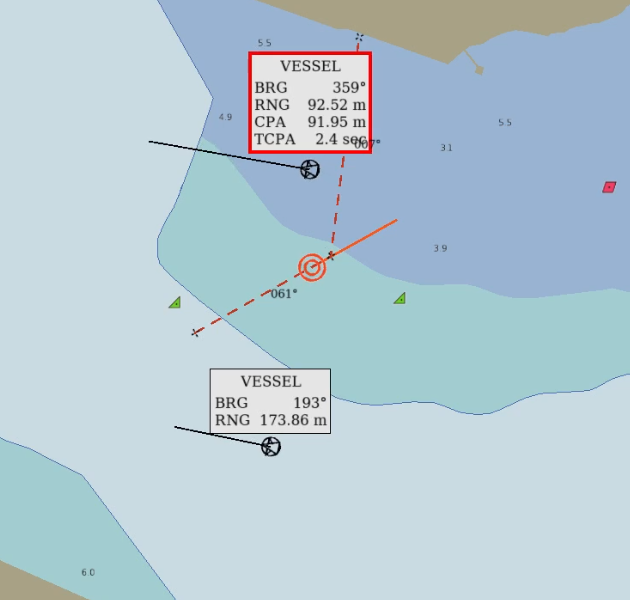}
	\caption{t=4m16s}\label{fig:t4m16s}
     \end{subfigure}
     \hfill
       \begin{subfigure}[b]{0.494\columnwidth}
        	\centering
        \includegraphics[width=0.95\textwidth]{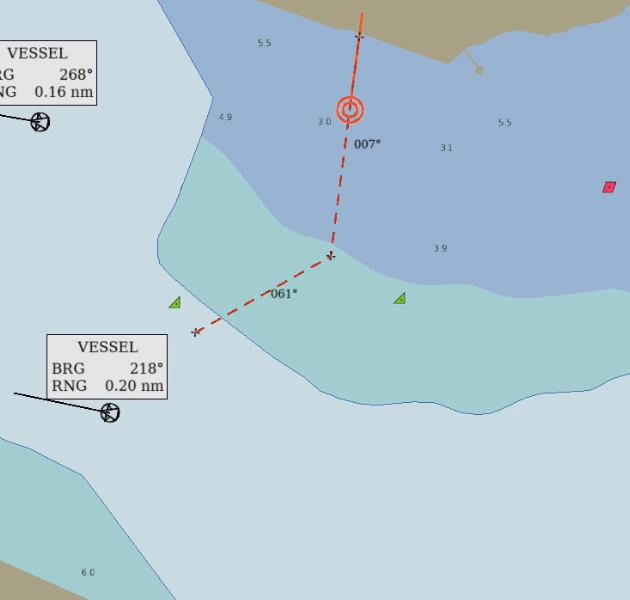}
	\caption{t=5m58s}\label{fig:t5m58s}
     \end{subfigure}
        \caption{Simulated validation of the SHP with most of the autonomy stack. The Greenhopper is departing the southern harbour, following the nominally planned path. A manoeuvre is required while underway, as two other vessels approach from starboard.}
        \label{fig:sas_shp_simulation}
\end{figure}
\section{Demonstration and discussion}
The autonomy stack and SHP was first validated using software-in-the-loop testing by simulating the sensor fusion output. The results are visualised on the HMI, and generated while running the MWS, SAS, ANS, ACS, RUT and SHP modules. The MWS acts as both a vessel and SFU simulator. Figure \ref{fig:sas_shp_simulation} shows a scenario in which the Greenhopper departs the southern harbour, following the nominally planned route. As the vessel is underway, two target vessels approach from the starboard side. The CPA for the southernmost target vessel is greater than the safety limit and is therefore not considered. The second target vessel violates the CPA limit. Therefore, once TCPA falls below a chosen limit, the SHP is triggered and a route deviation is calculated. Deviating allows the Greenhopper to safely avoid the target vessel before reaching its destination at the northern point of the fjord. Commissioning and full scale testing took place in the Fall of 2022.

A fundamental requirement for adhering to the COLREGs is an adequate situation assessment, as collision avoidance systems fail if the situation awareness is erroneous. MASS often depends on AIS for identifying other vessels, but as AIS is not compulsory for leisure crafts, it is crucial that the perception system can classify the perceived vessel as power-driven or not, and is furthermore able to determine whether a vessel is manoeuvrability restricted. Otherwise, the COLREGs cannot be applied correctly. For safe navigation, this is a major risk, since target vessels are commanded by humans, who expect that any vessel they encounter adhere to the COLREGs and knows how to mitigate the prevailing risk. 
If the scenario perceived by the SAS is correct, an appropriate collision avoidance strategy can be calculated by the SHP in a deterministic fashion within a finite time. If no solution exists during the voyage, the issue is raised to the ACS, which will call for assistance from the Remote Control Centre (RCC) or by stopping the vessel and signalling the surroundings that an emergency is unfolding.

\section{Conclusion}
The paper presented a collision avoidance perspective to autonomous ferries and harbour buses. A Danish autonomous ferry initiative, the Greenhopper, was introduced, and its autonomy stack was detailed and discussed. A deterministic collision avoidance strategy was presented, as well as simple ship domains for enforcing give-way responsibilities. The importance of a well-functioning and sufficiently accurate estimate of the unfolding situation was discussed in great detail. In conclusion, to navigate according to the COLREGs and safe navigation practises, the target vessels must be correctly classified in terms of vessel type and manoeuvrability.

Future work includes field verification of the proposed SHP in conjunction with the remaining autonomy stack. Once verified, the final steps towards fully commissioning the ferry for autonomous operation must be undertaken.

\begin{ack}
The authors acknowledge M. Bennedsen and J. B. Jensen from SIMAC for fruitful discussions on navigation, J. G. Mayer from Wärtsilä for data collection, and fellow ShippingLab contributors 
 for development of the autonomy stack. 
 This research was sponsored by the Danish Innovation Fund, The Danish Maritime Fund, Orients Fund, and the Lauritzen Foundation through the Autonomy part of the ShippingLab project, grant number 8090-00063B. The Danish Geodata Agency kindly provided the ENCs.
\end{ack}

\bibliography{ifacconf}             
                                                
\end{document}